\documentclass{l4dc2020} 

\usepackage{cleveref}
\usepackage{algorithm}
\usepackage{algorithmic}
\usepackage{amsmath}
\usepackage{graphics}
\usepackage{comment}

\renewcommand{\vec}[1]{\boldsymbol{#1}}
\newcommand{\mat}[1]{\boldsymbol{\mathrm{#1}}}
\newcommand{\trans}{^{\mkern-1.5mu\mathsf{T}}}

\crefname{figure}{Fig.}{Fig.}
\Crefname{figure}{Figure}{Figures}
\crefname{equation}{}{}
\Crefname{equation}{Equation}{Equations}

\title[Robust Online Model Adaptation]{Robust Online Model Adaptation by Extended Kalman Filter with Exponential Moving Average and Dynamic Multi-Epoch Strategy}
\usepackage{times}

\author{%
 \Name{Abulikemu Abuduweili} \Email{abduwali@pku.edu.cn}\\
 \addr School of Electronics Engineering and Computer Science, Peking University, Beijing 100871, P. R. China
 \AND
 \Name{Changliu Liu} \Email{cliu6@andrew.cmu.edu}\\
 \addr Robotics Institute, Carnegie Mellon University, Pittsburgh, PA 15213, USA%
}

\begin{document}

\maketitle

\vspace{-10pt}
\begin{abstract}%
High fidelity behavior prediction of intelligent agents is critical in many applications. 
However, the prediction model trained on the training set may not generalize to the testing set due to domain shift and time variance. 
The challenge motivates the adoption of online adaptation algorithms to update prediction models in real-time to improve the prediction performance.   
Inspired by Extended Kalman Filter (EKF), this paper introduces a series of online adaptation methods, which are applicable to neural network-based models. A base adaptation algorithm Modified EKF with forgetting factor (MEKF$_\lambda$) is introduced first, followed by exponential moving average filtering techniques. Then this paper introduces a dynamic multi-epoch update strategy to effectively utilize samples received in real time. With all these extensions, we propose a robust online adaptation algorithm: MEKF with Exponential Moving Average and Dynamic Multi-Epoch strategy (MEKF$_{\text{EMA-DME}}$). The proposed algorithm outperforms existing methods as demonstrated in experiments. The source code is open-sourced in the following link \url{https://github.com/intelligent-control-lab/MEKF_MAME}.
\end{abstract}


\vspace{-3pt}
\section{Introduction}\label{sec1}
\vspace{-3pt}
Supervised learning has been widely used to obtain models to predict the behaviors of intelligent agents \cite{rudenko2019human}. \textbf{Behavior  prediction}  is  a  sub-topic  of  time  series prediction \cite{weigend2018time}, which includes but is not limited to vehicle trajectory prediction during autonomous driving~\cite{lefevre2014survey} and human-motion prediction during human-robot collaboration~\cite{cheng2019human}.
Although a trained model typically performs well on the training set, performance can  drop significantly in a slightly different test domain or under a slightly different data distribution \cite{si2019agen,callison2010findings}. 
For tasks without annotated corpora from the test domain, adaptation techniques are required  to deal with the lack of domain-specific data.  
This paper studies robust online adaptation algorithms for behavior prediction.  

In \textbf{online adaptation}, a prediction model observes instances sequentially over time. After every observation, the model outputs a prediction and receives the ground truth. Then the online adaptation algorithm updates the prediction model according to the error measured between the prediction and the ground truth.  The goal of adaptation is to improve the prediction accuracy in subsequent rounds. An online adaptation algorithm is robust if it can efficiently adapt an existing model to a different (test) data distribution, without generating big transient errors.

For prediction models encoded in neural networks, most existing online adaptation approaches are based on stochastic gradients \cite{kivinen2004online}. For example, the identification-based approach uses stochastic gradient descent (SGD) to adapt the model online \cite{bhasin2012robust}.  
However, these methods may be sub-optimal in minimizing the local prediction errors. Another solution is to use the recursive least square parameter adaptation algorithm (RLS-PAA) \cite{Ljung2010A}, which has been applied to adapt the last layer of a feedforward neural network \cite{cheng2019human} or the last layer of a recurrent neural network \cite{si2019agen}. RLS-PAA can only adapt the last layer of a neural network since it only applies to linear models. 
To adapt other layers, the adaptation problem becomes nonlinear, which requires the development of robust optimal nonlinear adaptation algorithms~\cite{jazwinski2007stochastic,cooper2014use,abuduweili2019adaptable}.  

Since a neural network parameterizes a nonlinear system with a layered structure, learning or adaptation of the neural network is equivalent to parameter estimation of the nonlinear system. 
The extended Kalman filter (EKF) is one promising method for nonlinear parameter estimation \cite{jazwinski2007stochastic}, which is derived by linearizing the system equations at each time step and applying Kalman filter (an optimal filter that minimizes the tracking error) on the linearized system. The EKF approach has been demonstrated to be superior to the SGD-based algorithms in training feedforward neural networks \cite{iiguni1992real,ruck1992comparative}. 
Nonetheless, in online adaptation, more recent data is more important \cite{Fink2001Non}. Similar to adaptive EKF methods \cite{yang2006adaptive,ozbek2004adaptive,anderson2012optimal} that discount old measurements, this paper considers the Modified Extended Kalman Filter with forgetting factor, MEKF$_\lambda$, as a base adaptation algorithm. 

On top of the base adaptation algorithm, the following modifications are made. 
Generally, the step size of parameter update in EKF-based approaches may not be optimal, due to the error introduced during linearization. Inspired by \textbf{exponential moving average (EMA)} methods,  this paper proposes EMA filtering to the base MEKF$_\lambda$ in order to increase the convergence rate. The resulting algorithm is called MEKF$_{\rm EMA}$. 
Then in order to effectively utilize the samples in online adaptation, this paper proposes a \textbf{dynamic multi-epoch update} strategy to  discriminate the ``hard" samples from ``easy" samples, and sets different weights for them. The dynamic multi-epoch update strategy can improve the effectiveness of online adaptation with any base optimizers, e.g., SGD or MEKF$_{\rm EMA}$. By incorporating MEKF$_{\rm EMA}$ with the dynamic multi-epoch update strategy, we propose the algorithm MEKF$_{\text{EMA-DME}}$ (MEKF with Exponential Moving Average and Dynamic Multi-Epoch update strategy). 

The remainder of the paper first formulates the online adaptation problem, then discusses the proposed algorithm, and finally validates the effectiveness and flexibility of the proposed algorithms. 

\vspace{-10pt}
\section{Online adaptation framework}\label{sec2}
\vspace{-3pt}
The behavior prediction problem is to make inference on the future behavior of the target agent given the past and current measurement of the target agent and its surrounding environment. The transition model for behavior prediction problem is formulated as \vspace{-3pt}
\begin{equation}\small
\mat{Y}_t = f(\mat{\theta}, \mat{X}_t),\vspace{-7pt} \label{eq:traj_pred}
\end{equation}
where the input vector $\mat{X}_t=[\Vec{x_t}; \Vec{x_{t-1}}; \cdots;\Vec{x_{t-n+1}}]$ denotes the stack of $n$-step current and past measurements (e.g. trajectory of states or extracted features) at time steps $t, t-1, \ldots, t-n+1$.   The output vector $\mat{Y}_t=[\Vec{y_{t+1}}; \Vec{y_{t+2}}; \cdots; \Vec{y_{t+m}}]$ denotes the stack of the $m$-step future behavior (e.g. future trajectory) at time steps $t+1, t+2, \ldots, t+m$. The function $f$ is the prediction model that maps the measurements to the future behavior. $\mat{\theta}$ denotes the (ground truth) parameter of the model. It is assumed that there are recurrent structures in $f$ such that the prediction of $\mat{Y}_t$ is made by rolling out the one-step predictions $\Vec{y_{t+1}} = f_1(\mat{\theta}, \mat{X}_t)$. The function $f_1$ is the one-step prediction function and is a recurrent part of the overall prediction model.

Online adaptation explores local overfitting to minimize the prediction error. At time step $t$, the following prediction error is to be minimized \vspace{-3pt}
\begin{equation}\small
\min_{\mat{\hat{\theta}}_t} \|\mat{Y}_t - f(\mat{\hat{\theta}}_t, \mat{X}_t)\|_p,
\vspace{-7pt}
\end{equation}
where $\mat{Y}_t$ is the ground truth trajectory (to be observed in the future) and $\mat{\hat{Y}}_t:= f(\mat{\hat{\theta}}_t, \mat{X}_t)$ is the predicted trajectory using the estimated model parameter $\mat{\hat{\theta}}_t$. The adaptation objective can be in any $\ell_p$ norm. This paper considers  $\ell_2$ norm. Assume that the true model parameter changes slowly during adaptation, i.e., $\mat{\dot\theta} \approx 0$. Then the estimated model parameter that minimizes the prediction error in the future can be approximated by the estimated parameter that minimizes the fitting error in the past. Solving for the estimated parameter that minimizes the fitting error corresponds to a nonlinear least square (NLS) problem.

\vspace{-7pt}
\begin{definition}[Problem NLS]
Given a dataset $\{(\mat{X}_i,\mat{Y}_i),i=1,2,\cdots,T\}$, find $\mat{\hat{\theta}}_t \in {\mathbb{R}}^n$ that minimizes  $J_t(\mat{\hat{\theta}}_t) = \frac{1}{t} \sum_{i=1}^t{\| \mat{e}_i \|_2^2}$, 
where error term is defined as $\mat{e}_i =\mat{Y}_i-f(\mat{\hat{\theta}}_t, \mat{X}_i) $. \label{nls}
\end{definition}
\vspace{-7pt}

In online adaptation, the estimate of the model parameter is updated iteratively when new data is received. Then a new prediction is made using the new estimate. In the next time step, the estimate will be updated again given the new observation and the process repeats. It is worth noting that the observation we received at time $t$ is $\vec{y_{t}}$. The other terms in $\mat{Y_t}$ remain unknown. This paper is focused on adaptation methods using only one-step observation. It is possible to adapt with multi-step observations, which will be studied in the future.  
The process for online adaptation  is summarized in \cref{algo:online_adapt}. $\mat{\hat{\theta}}_t$ is the estimate of the model parameter $\mat{\theta}$ at time $t$. 

\begin{algorithm}\small
\caption{Generic Online Adaptation (Adaptable Prediction) \label{algo:online_adapt}}
\begin{algorithmic}[1]
\REQUIRE Initial model parameters $\mat{\hat{\theta}_0}$
\ENSURE sequence of predictions $\{\mat{\hat{Y}}_t \}_{t=1}^T$
\FOR { $t = 1,2, \cdots,T $} 
\STATE obtain the ground truth observation value $\vec{y_{t}}$; construct the input features $\mat{X}_t$ 
\STATE adaptation~step~(supervised): ~ $\mat{\hat{\theta}}_{t} = {\rm Adapt}(\mat{\hat{\theta}}_{t-1}, \vec{\hat{y}_{t}},\vec {y_{t}}) $ \label{algo:adapt_step}
\STATE prediction~step: ~ $\mat{\hat{Y}}_t =f(\mat{\hat{\theta}}_t, \mat{X}_t)$, where $\mat{\hat{Y}}_t = [\vec{\hat{y}_{t+1}};\vec{\hat{y}_{t+2}};\cdots;\vec{\hat{y}_{t+m}}] $
\ENDFOR 
\RETURN sequence of predictions $\{\mat{\hat{Y}}_t \}_{t=1}^T$
\end{algorithmic}
\end{algorithm}

\vspace{-20pt}
\section{Robust nonlinear adaptation algorithms}\label{sec3}
\vspace{-3pt}

\subsection{Modified EKF with forgetting factor}
\vspace{-3pt}
Our base adaptation algorithm is inspired by the recursive EKF method \cite{moriyama2003incremental,alessandri2007recursive}.
In EKF, the object being estimated is the state value of a dynamic system, while in adaptable prediction, the object to be adapted is the parameters that describe the system dynamics. Nonetheless, we can apply the EKF approach to adapt model parameters by regarding model parameters as system states. 
By assuming that the ground truth $\mat{\theta}$ changes very slowly, we can pose the parameter adaptation problem as a static state estimation problem \cite{ruck1992comparative,nelson2000nonlinear} with the following dynamics,\vspace{-3pt}
\begin{eqnarray}\small
\vspace{-5pt}
\mat{\hat{\theta}}_{t} &=& \mat{\hat{\theta}}_{t-1} + \mat{
\omega}_{t}, \\
\vec{y_{t}} &=& f_1(\mat{\hat{\theta}}_{t-1},\mat{X}_{t-1}) +  \mat{u}_t, 
\vspace{-7pt}
\end{eqnarray}
where $\mat{\hat{\theta}}_t$ is an estimate of the (ground truth) model  parameter $\mat{\theta}$; $\vec{y_{t}}$ is the observation at time $t$; and $\vec{\hat{y}_{t}}=f_1(\mat{\hat{\theta}}_{t-1},\mat{X}_{t-1})$ is the prediction for time step $t$ made at time $t-1$. $f_1$ is the one-step prediction function. The injected process noise $\mat{\omega}_t \sim \mathcal{N}(0,\mat{Q}_t)$ and the injected measurement noise $ \mat{u}_t \sim \mathcal{N}(0,\mat{R}_t)$ are assumed to be zero mean Gaussian white noise, and are identically and independently distributed. The symbol $\mathcal{N}$ represents Gaussian distribution. $\mat{Q}_t$ and $\mat{R}_t$ represent the covariance matrices for the process noise and the measurement noise respectively, which should be positive semidefinite. If there is no knowledge about the cross correlation of the  noises, it is reasonable to assume that the entries in the noise vector are independent of each other, and set $\mat{Q}_t$ and $\mat{R}_t$ to be proportional to the identity matrix. For simplicity, we assume $\mat{Q}_t = \sigma_q \mat{I}$ and $\mat{R}_t = \sigma_r \mat{I}$ for $\sigma_q \ge 0$ and $\sigma_r>0$.

In online adaptation, we assume that data in the
distant past is no longer relevant for modeling the current dynamics, i.e. more recent data is more important. Hence, 
we consider a weighted nonlinear recursive least squares (NLS) problem:\vspace{-5pt}
\begin{align}\small
\vspace{-5pt}
\min_{\mat{\hat{\theta}}_t}  \sum_{i=1}^t{\lambda^{t-i}\|\vec{y_{i}} - f_1(\mat{\hat{\theta}}_{i-1},\mat{X}_{i-1})\|_2^2}, \qquad 0 < \lambda \le 1, \label{eq:nls}
\vspace{-5pt}
\end{align}
where $\lambda$ is the ``forgetting factor" which provides exponential decay to older samples. 
The forgetting factor prevents the EKF from saturation, and increases the algorithm's ability to track a changing system. \Cref{algo:mekf_lambda} summarizes the modified extended Kalman filter algorithm  with forgetting factor (\textbf{ MEKF$_\lambda$}).

\begin{algorithm}\small
\caption{Modified EKF algorithm  with forgetting factor (${\rm MEKF}_\lambda$) \label{algo:mekf_lambda}}
\begin{algorithmic}[1]
\REQUIRE{Initial hyper-parameter for ${\rm MEKF}_\lambda$:  $p_0 > 0 ,~ \lambda > 0  ,~ \sigma_r > 0, ~ \sigma_q \ge 0;  ~ \mat{P}_0=p_0\mat{I}$ }
\REQUIRE{previous parameter $\mat{\hat{\theta}}_{t-1}$, previous prediction $\vec{\hat{y}_{t}}=f_1(\mat{\hat{\theta}}_{t-1}, \mat{X}_{t-1})$, current observation $\vec{y_{t}}$ at time step $t$}
\ENSURE Adapted parameter $\mat{\hat{\theta}_{t}}$
\STATE $\mat{H}_t = \frac{\partial f_1(\mat{\theta},\mat{X})}{ \partial \mat{\theta}} \mid_{\mat{\theta}=\mat{\hat{\theta}}_{t-1},\mat{X}=\mat{X}_{t-1}}$
\STATE $\mat{K}_t = \mat{P}_{t-1} \cdot \mat{H}_t \trans \cdot (\mat{H}_t \cdot \mat{P}_{t-1} \cdot \mat{H}_t \trans + \sigma_r \mat{I})^{-1}$
\STATE $\mat{\hat{\theta}}_{t} =\mat{\hat{\theta}}_{t-1} + \mat{K}_t \cdot (\vec{y_t} - f_1(\mat{\hat{\theta}}_{t-1},\mat{X}_{t-1})) \label{algo:ekf_w}$
\STATE $ \mat{P}_{t} = \lambda^{-1}(\mat{P}_{t-1} - \mat{K}_t\cdot \mat{H}_t \cdot \mat{P}_{t-1} +  \sigma_q \mat{I}) $
\end{algorithmic}
\end{algorithm}

In \cref{algo:mekf_lambda}, $\mat{K}_t$ is the Kalman gain.  $\mat{P}_t$ is a matrix representing the uncertainty in the estimates of the model parameter $\mat{\theta}$. $\mat{H}_t$ is the
gradient matrix by linearizing the network. In online adaptation, $\mat{\theta}_0$ is initialized by the offline trained parameter of the model. For $\mat{P}_0$, due to the absence of  a priori information, the $\mat{P}_0$ matrix can be set to be proportional to the identity matrix, i.e. $\mat{P}_0 = p_0 \mat{I}$ for $p_0>0$.  

\vspace{-5pt}
\subsection{Extensions with exponential moving average filtering}\label{sec:3extension}
\vspace{-3pt}
In the following discussion, for simplicity, an optimizer (e.g. MEKF$_\lambda$) that solves the adaptation problem will be denoted as $A_{\mat{P}}$ with internal state matrix $\mat{P}$.  The optimization process for adaptation can be compactly written as\vspace{-10pt}
\begin{align}\small
\vspace{-5pt}
\begin{split}
    \mat{\hat{\theta}}_t &= \mat{\hat{\theta}}_{t-1} + \mat{V}_{t}, \\
    \mat{V}_{t} &= A_{\mat{P}}(\mat{\hat{\theta}}_{t-1},\vec{\hat{y}_{t}},\vec{y_{t}} ).
\end{split} \label{eq:optim_process}
\vspace{-7pt}
\end{align}
where 
$\mat{V}_{t}$ is the step size of the parameter update at time step $t$. 

To speed up the convergence of parameter estimation, we propose to apply exponentially-decayed moving average (EMA) for filtering in the MEKF$_\lambda$ optimization process. In SGD-based methods, 
numerous variants of EMA have been successfully used to speed up the convergence, including Polyak averaging \cite{polyak1964some} and momentum \cite{qian1999momentum}. 
For MEKF$_\lambda$, we can either
apply EMA on the step size $\mat{V}$, to be discussed as EMA-V or momentum; or apply EMA on the optimizer's inner state $\mat{P}$, to be discussed as EMA-P. 

\paragraph{EMA-V}
EMA-V or momentum is widely used in SGD-based optimization algorithms \cite{qian1999momentum},
which helps accelerate gradient-based optimizers in relevant directions and dampen oscillations \cite{qian1999momentum}. Momentum can be regarded as an EMA filter on the step size of parameter update. 
It calculates the step size $\mat{V}_{t}$ by decreasing exponentially the older step size with a factor $\mu_v\in[0,1]$, i.e. $\mat{V}_{t} = \mu_v \mat{V}_{t-1} +(1 -  \mu_v)  A_{\mat{P}}(\mat{\hat{\theta}}_{t-1},\vec{\hat{y}_{t}},\vec{y_{t}} )$.  

\paragraph{EMA-P \label{sec:ema_p}}
As mentioned earlier in MEKF$_\lambda,~\mat{P}_t$ is a matrix representing the uncertainty in the parameter estimates. In order to attenuate instability during adaptation caused by anomaly data, we can smooth the inner state of the optimizer by pre-filtering $\mat{P}_t$. The principle of pre-filtering the inner state (e.g., gradient, adaptive learning rate) before using them in optimization is applicable to many optimization algorithms. For example, in Adam \cite{kingma2014adam}, the  estimate  of the first and second moment is filtered every step using EMA. 
Similarly, we can apply EMA on the inner state matrix $\mat{P}_t$ of ${\rm MEKF}_\lambda$.

By combining EMA-V and EMA-P, we propose the modified extended Kalman  filter with  exponential moving average (\textbf{ MEKF$_{\rm EMA}$}) algorithm as shown in \cref{algo:ekfema}, where $\mu_v$ is a momentum factor, and
$\mu_p$ is a decay factor for the EMA filtering of $\mat{P}_t$. 

\begin{algorithm}\small
\caption{Modified Extended Kalman Filter with  Exponential Moving Average Filtering \label{algo:ekfema}}
\begin{algorithmic}[1]
\REQUIRE{Initial base hyper-parameter:  $p_0 > 0 ,~ \lambda > 0  ,~ \sigma_r > 0, ~ \sigma_q \ge 0 ; \mat{P}_0 = p_0\mat{I}$}
\REQUIRE{Initial EMA hyper-parameter:  $0 \le \mu_v < 1 ,~ 0 \le \mu_p < 1$}
\REQUIRE{previous parameter $\mat{\hat{\theta}}_{t-1}$, previous prediction $\vec{\hat{y}_{t}}=f_1(\mat{\hat{\theta}}_{t-1}, \mat{X}_{t-1})$, current observation $\vec{y_{t}}$ at time step $t$}
\ENSURE {Adapted parameter $\mat{\hat{\theta}}_t$}
\STATE $\mat{H}_t = \frac{\partial f_1(\mat{\theta},\mat{X})}{ \partial \mat{\theta}} \mid_{\mat{\theta}=\mat{\hat{\theta}}_{t-1},\mat{X}=\mat{X}_{t-1}} 
\label{algo:ekfema_h}$
\STATE $\mat{K}_t = \mat{P}_{t-1} \cdot \mat{H}_t \trans \cdot (\mat{H}_t \cdot \mat{P}_{t-1} \cdot \mat{H}_t \trans + \sigma_r \mat{I})^{-1} $
\STATE $\mat{V}_{t} = \mu_v \mat{V}_{t-1} +(1 -  \mu_v)   \mat{K}_t \cdot (\vec{y_t} - f_1(\mat{\hat{\theta}}_{t-1},\mat{X}_{t-1})) 
\label{algo:ekfema_v}$
\STATE $ \mat{\hat{\theta}}_{t} = \mat{\hat{\theta}}_{t-1} + \mat{V}_{t} $ 
\STATE $\mat{P}_{t}^\ast = \lambda^{-1}(\mat{P}_{t-1} - \mat{K}_t\cdot \mat{H}_t \cdot \mat{P}_{t-1} +  \sigma_q \mat{I})$
\STATE $ \mat{P}_{t} =\mu_p \mat{P}_{t-1} + (1 - \mu_p) \mat{P}_{t}^\ast
 $ \label{algo:eq_gema} 
\end{algorithmic}
\end{algorithm}

\vspace{-10pt}
\subsection{Dynamic multi-epoch update strategy}\label{sec:multi_epoch}
\vspace{-3pt}
In generic online adaptation, all data are equally considered. We run the adaptation algorithm chronologically from the first data $\mat{X}_1$ to the last data $\mat{X}_T$. Every data sample is used only once, as shown in \cref{algo:online_adapt}. We call the adaptation method that uses every data sample only once as \textbf{single-epoch online update strategy}. 

Inspired by curriculum learning \cite{bengio2009curriculum} in offline training, we introduce a more effective way to determine the adaptation epochs for every data sample during online adaptation. A curriculum can be viewed as a sequence of training criteria. Each training criterion in the sequence is associated with a different set of weights on the training examples. 
That said, it is practically useful to differentiate ``easy" samples from ``hard" samples. In the online adaptation scenario, 
we introduce the following dynamic multi-epoch strategy to mimic curriculum learning.

\vspace{-7pt}
\begin{definition}[Dynamic multi-epoch online update strategy]
In online adaptation, 
the predicted output $\vec{\hat{y}_{t}}$ generated by the estimated parameter $\mat{\theta^\ast}$ is $\vec{\hat{y}_{t}} = f_1(\mat{\theta^\ast},\mat{X}_{t-1})$. 
Define a criterion $\mathcal{C}$ to determine the number of epochs $\kappa_{t}~(\kappa_{t} \in \mathbb{N})$ to adapt the parameter with the  current sample, i.e., $\kappa_{t} = \mathcal{C}(\mat{X}_{t-1},\vec{y_{t}},\vec{\hat{y}_{t}},\mat{\theta^\ast}) $. In other words, we reuse  the input-output pair $(\mat{X}_{t-1}, \vec{y_{t}})$ $\kappa_{t}$ times to adapt the parameter $\mat{\theta^\ast}$. 
This approach is called the dynamic multi-epoch online update strategy or \textbf{dynamic multi-epoch update}.
\end{definition}
\vspace{-7pt}

We propose a very straighforward criterion $\mathcal{C}$ to determine the number of epochs $\kappa_t$ for every sample, as shown in \cref{algo:multi_epoch}\footnote{The criterion can be application-specific. This criterion is proposed since it is observed in our experiment that a large number of epochs will lead to overfitting to the historical data, which is harmful to generalization. We will investigate more reasonable and effective criterion for dynamic multi-epoch update in the future.}. Two thresholds $\xi_1$ and $\xi_2$ are used to discriminate ``easy", ``hard", and ``anomaly" samples. Before updating the parameter, we calculate the prediction error $j_{t} = \|\vec{y_{t}} - \vec{\hat{y}_{t}} \|_2$ at the current step. If the error satisfies $j_{t} < \xi_1 $, the sample is considered as an  ``easy" sample. Then we run single-epoch update for this sample. If the error satisfies $\xi_1 \le j_{t} < \xi_2 $, the sample is considered as a ``hard" sample. Then we reuse this sample and run the adaptation twice. The rationale is that for a ``hard" sample, an adaptation optimizer may not learn  enough under single-epoch update.  If the error satisfies $ j_{t} \ge \xi_2$, the sample is considered as an  ``anomaly" sample. Then we skip the update of $\mat{\hat{\theta}}_t$. The rationale is that if the cost is too high, the sample is likely to be an anomaly point in the data distribution, which may destabilize the model adaptation process if learned. 
It is crucial to identify and learn more from those ``hard" samples without sacrificing the generalizability of the model.

\begin{algorithm}\small
\caption{Dynamic Multi-epoch Update Strategy  \label{algo:multi_epoch}}
\begin{algorithmic}[1]
\REQUIRE{threshold $0 \le \xi_1 \le \xi_2$, optimizer $A_{\mat{P}}$ (e.g. ${\rm MEKF}_{\rm EMA}$ ), model prediction function $f_1$}
\REQUIRE{previous parameter $\mat{\hat{\theta}}_{t-1}$, previous input $\mat{X}_{t-1}$, previous prediction $\vec{\hat{y}_{t}}=f_1(\mat{\hat{\theta}}_{t-1}, \mat{X}_{t-1})$,  current observation $\vec{y_{t}}$}
\ENSURE Adapted parameter $\mat{\hat{\theta}}_t$
\STATE $j_{t} = \|\vec{y_{t}} - \vec{\hat{y}_{t}} \|_2$   
\STATE $ \kappa_{t} = 
\begin{cases}
1 \ , \ \textbf{If} ~ j_{t} < \xi_1 \\
2 \ , \  \textbf{If} ~ \xi_1 \le j_{t} < \xi_2 \\
0 \ , \ \textbf{If} ~ j_{t} \ge \xi_2 \\
\end{cases}$
\STATE $\mat{\theta}_{t,0}^\ast = \mat{\hat{\theta}}_{t-1}$
\FOR{ $i = 1, \cdots,\kappa_t $} 
\STATE $\vec{\hat{y}^\ast_{t}}=f_1(\mat{\theta}_{t,i-1}^\ast, \mat{X}_{t-1}) $
\STATE $\mat{\theta}_{t,i}^\ast = \mat{\theta}_{t,i-1}^\ast + A_{\mat{P}}(\mat{\theta}_{t,i-1}^\ast, \vec{\hat{y}^\ast_{t}},\vec {y_{t}}) $
\ENDFOR
\STATE $\mat{\hat{\theta}}_{t} = \mat{\theta}_{t,\kappa_t}^\ast$
\end{algorithmic}
\end{algorithm}

The thresholds $\xi_1$ and $\xi_2$ can be determined by the validation set empirically. If the dataset is noise-free, there is no need to identify ``anomaly" samples and we set  $\xi_2 \to + \infty$. In general, we recommend the following method to find  the desired $\xi_1$ and $\xi_2$. First, we need to run the single-epoch adaptation on the validation set and record each sample's prediction error $\{j_1,j_2,\cdots\}$. Second, we set $\xi_1$ as the $50\% \sim 95\%$ quantile value of the errors, and set  $\xi_2$ as the $99.9\% \sim 100\%$ quantile value of the errors. That means, we regard  $50\% \sim 95\%$ of the samples as ``easy" samples,  $5\% \sim 50\%$ of the samples as ``hard" samples, and  $0\% \sim 0.1\%$ of the samples as ``anomaly" samples.

We use  \textbf{MEKF$_{\text{EMA-DME}}$} to denote MEKF$_{\rm EMA}$ with the dynamic multi-epoch update strategy. 

\vspace{-3pt}
\section{Experiments}\label{sec4}
\vspace{-3pt}
\paragraph{Experimental design}
In the experiments, we consider multi-task prediction problems for simultaneous intention and trajectory prediction of either humans or vehicles.
We construct  Recurrent Neural Network \cite{salehinejad2017recent} (RNN) based architectures to conduct experiments on Mocap dataset (human) and NGSIM dataset (vehicle) \cite{colyar2007us}. We leaved the details of the experiment in the  \Cref{ap:design}. 
Before online adaptation, the prediction models are trained offline. In the following discussion, we studied the performance of various adaptation algorithms on these offline-trained models (with online adaptation on the test set). In particular, we evaluate the accuracy (0-1) for intention prediction, and the mean squared error (MSE) for trajectory prediction.


\paragraph{Comparison among different optimizers\label{sec:optimizer}}
The proposed algorithm MEKF$_{\text{EMA-DME}}$ is compared with the based algorithm MEKF$_\lambda$ and other commonly used optimizers such as SGD, Adam, and Amsgrad. For fair comparison, we apply the dynamic multi-epoch update strategy on SGD, Adam, and Amsgrad.

\begin{table}[!htbp]
\vspace{-10pt}
\centering
\caption{Comparison of different optimizers.  }
\label{tab:optim_mocap}\scriptsize
\begin{tabular}{l|l|l|l|l|l|l|l}
\hline \hline
Dataset & Metrics & w/o adapt & SGD & Adam & Amsgrad & MEKF$_\lambda$ &MEKF$_{\text{EMA-DME}}$ \\ \hline
{Mocap} & accuracy & 0.984 & 0.984 & 0.984 & 0.984 & 0.985 & \textbf{0.985} \\
 & MSE(dm$^2$) & 3.271 & 3.185 & 3.149 & 3.156 & 2.788 & \textbf{2.746} \\ \hline
{NGSIM} & accuracy & 0.951 & 0.954 (.0427) & 0.951 (.0426) & 0.951 (.0427) & 0.956 (.0430) & \textbf{0.956} (.0430) \\
 & MSE(m$^2$) & 2.559 & 2.367 (1.981) & 2.402 (1.975) & 2.407 (1.993) & 2.157 (1.902) & \textbf{2.092} (1.893) \\ \hline \hline
\end{tabular}
\vspace{-10pt}
\end{table}

\Cref{tab:optim_mocap} shows the prediction performance of online adapted models using different optimizers on the Mocap dataset and the NGSIM dataset. 
Compared to the stochastic gradient-based algorithms, the EKF-based methods MEKF$_\lambda$ and MEKF$_{\text{EMA-DME}}$ perform better. 
In addition, MEKF$_{\text{EMA-DME}}$ has the best performance among all, due to the extensions inspired by EMA and dynamic multi-epoch update. On the CMU Mocap dataset,  Adam reduces the trajectory MSE by 3.73\%. MEKF$_\lambda$ reduces the trajectory MSE by 14.77\%. MEKF$_{\text{EMA-DME}}$ reduces the trajectory MSE by 16.05\%. These improvements are important to ensure safe and efficient operation of behavior prediction \cite{zhao2020experimental}. 
The variance of perfomance using different optimizers  on   the  NGSIM  dataset is shown in the parenthesis. In particular, the standard deviation (Std.) of the prediction accuracy as well the Std. of MSE are shown. For intention prediction, MEKF$_\lambda$ and MEKF$_{\text{EMA-DME}}$ have slightly higher Std. than other SGD based optimizers.   However, for trajectory prediction,  MEKF$_\lambda$ and MEKF$_{\text{EMA-DME}}$ have lower Std. than other SGD based optimizers. 

The running time of the adaptation algorithm correlates with the number of parameters to adapt. For the time complexity analysis,  SGD, Adam, and Amsgrad have similar complexity, while MEKF$_\lambda$ and MEKF$_{\rm EMA}$ have similar complexity. So we only compare MEKF$_\lambda$ with  SGD below. For adapting encoder’s
hidden layers (12480 parameters), SGD takes  0.08 seconds per sample and MEKF$_\lambda$ takes 0.41 seconds per sample\footnote{The running time is evaluated on the NGSIM dataset using a Ubuntu desktop with Intel Core i9-9940X CPU (3.30GHz) and GeForce RTX 2080 Ti GPU. }. For real-time adaptable prediction, as long as the number of parameters to adapt is not too big, MEKF$_\lambda$ can meet the real-time requirements.

\paragraph{Effectiveness of extensions\label{sec:extension}}

This section studies the effectiveness of the proposed extensions to MEKF$_\lambda$ in \cref{sec:3extension,sec:multi_epoch}. The hyperparameters are set as $\mu_p=0.3$ and  $\mu_v=0.3$. The results are shown in \cref{tab:extension_mocap}. 

\begin{enumerate}
 \vspace{-7pt}
\setlength\itemsep{-0.2em}
\item EMA-V or momentum barely improves the performance.
Two potential reasons are:
1) The momentum does not help EKF-based optimizers. In every optimization step, EKF-based optimizers has already incorporated the historical data. Hence its step size $\mat{V}_t$ is already closer to optimum than that of SGD. The learning gain in SGD is not based on historical data but manually defined. 
2) The moving average on the parameter or the step size is more applicable to offline training than to online adaptation. The inapplicability is due to the fact that online adaptation can only process data sequentially in time, which is significantly different from the shuffled, repetitive, and batched process in offline training.  
\item EMA-P slightly improves the performance of MEKF$_\lambda$. Filtering of $\mat{P}_t$ can smooth the inner state and improve convergence.
\item Dynamic multi-epoch update improves the performance of MEKF$_\lambda$, and it has the best performance among all the proposed extensions.
\vspace{-7pt}
\end{enumerate}

We design the additional experiment about different criteria for DME  in \Cref{ap:dme}  
to compared the proposed criteria (Under the error spectrum,
the first 50\% are ”easy” samples, the middle 50\% to 99.9\% are ”hard” samples, and the last
0.1\% are ”anomaly” samples.) with fixed 2-epoch criteria (Each sample was used  twice.) and random criteria (Which has same ”easy” and ”hard” ratio as the proposed criterion, but distinguishing ”easy”,
”hard” and ”anomaly” samples randomly.). The results in \Cref{ap:dme} show that: the proposed criterion outperforms other criteria, which justifies
the effectiveness of the proposed error-based criterion. Nonetheless, we will investigate more
reasonable and effective criterion for dynamic multi-epoch update in the future.

\begin{table}[!htbp]
\vspace{-10pt}
\centering
\caption{Performance of MEKF$_\lambda$ extensions.}
\label{tab:extension_mocap}\footnotesize
\begin{tabular}{l|l|l|l|l|l}
\hline \hline
Dataset & Metrics & MEKF$_\lambda$ & MEKF$_\lambda$ + EMA-V & MEKF$_\lambda$ + EMA-P & MEKF$_\lambda$ + DME \\ \hline
Mocap & accuracy & 0.985 & 0.985 & 0.985 & 0.985 \\
 & MSE(dm$^2$) & 2.788 & 2.790 & 2.775 & 2.749 \\ \hline
NGSIM & accuracy & 0.956 & 0.956 & 0.956 & 0.956 \\
 & MSE(m$^2$) & 2.157 & 2.156 & 2.123 & 2.122 \\ \hline \hline
\end{tabular}
\vspace{-10pt}
\end{table}

\vspace{-3pt}
\section{Conclusions}\label{sec5}
\vspace{-3pt}
This paper studied online adaptation of neural network-based prediction models for behavior prediction. An EKF-based adaptation algorithm  MEKF$_\lambda$ was introduced as an effective base algorithm for online adaptation. In order to improve the performance and convergence of MEKF$_\lambda$,  exponential moving average filtering was investigated, including momentum and EMA-P. Then this paper introduced a dynamic multi-epoch update strategy, which is compatible with any optimizer. By combining all extensions with the base MEKF$_\lambda$ algorithm, we introduced the robust online adaptation algorithm MEKF$_{\text{EMA-DME}}$. In the experiments, we demonstrated the effectiveness of the proposed adaptation algorithms. 

In the future, mathematical analysis of the proposed online adaptation algorithm MEKF$_{\text{EMA-DME}}$ will be performed in order to provide theoretical guarantees on stability, convergence, and boundedness. In addition, we will apply the proposed  algorithm on a wider range of problems in addition to behavior prediction problems.

\bibliography{MEKF_EMA_DME}

\appendix
\section{Detailed Experiments} \label{ap}
 
\subsection{Experimental design} \label{ap:design}

\subsubsection{Multi-task prediction}
In experiment we considers a multi-task prediction problem for simultaneous intention and trajectory prediction.  Intentions are discrete representations of future trajectories. For example, in vehicle behavior prediction, intention can be acceleration and deceleration in a certain time window in the future.

The transition models for trajectory and intention prediction of the target agent are formulated as 
\begin{align}
 \vspace{-5pt}
\mat{Y}_t &= f(\mat{\theta}, \mat{X}_t)\\
\Vec{z_t} &= g(\mat{\theta}, \mat{X}_t)
 \vspace{-5pt}
\end{align}
where the input vector $\mat{X}_t$ denotes the stack of $n$-step current and past measurement at time steps $t, t-1, \ldots, t-n+1$. The measurement $\Vec{x_t}$ can include the position and velocity of the target agent as well as the state of the environment. For human behavior prediction, this paper uses raw position and velocity measurements of the human.  For vehicle behavior prediction, this paper additionally uses environment features. The output vector $\mat{Y}_t$ denotes the stack of the $m$-step future trajectory at time steps $t+1, t+2, \ldots, t+m$. Another output vector $\vec{z_t}$ is a probability distribution over different intentions at time step $t$. The function $f$ maps current and past measurements to the future trajectory, while the function $g$ maps current and past measurements to the current intention.

One possible design of the multi-task prediction model is to use an encoder-decoder-classifier architecture. The encoder serves as a common part for all sub-tasks, which maps the input vector $\mat{X}_t$ to a hidden representation $\mat{h}_t$. The decoder works for trajectory prediction, which maps the hidden representation $\mat{h}_t$ to the predicted future trajectory $\mat{Y}_t$. The classifier aims to predict the intention $\vec{z_t}$ from the hidden representation $\mat{h}_t$. Mathematically, the relationships among the encoder, the decoder, and the classifier are:
\begin{align}
 \vspace{-5pt}
\mat{h}_t &= \rm{Encoder}(\mat{\theta^E}, \mat{X}_t), \\
\mat{Y}_t &= \rm{Decoder}(\mat{\theta^D}, \mat{h}_t), \\
\mat{z_t} &= \rm{Classifier}(\mat{\theta^C}, \mat{h}_t),
 \vspace{-5pt}
\end{align}
where $\mat{\theta^E}$ is the parameter for the encoder, which affects  all sub-tasks, $\mat{\theta^D}$ is the parameter for the decoder, and $\mat{\theta^C}$ is the parameter for the classifier. The total (ground truth) parameter of the model is $\mat{\theta}:= \{\mat{\theta^E},\mat{\theta^D},\mat{\theta^C}\}$.

\begin{figure}[htbp]
\centering
\includegraphics[width=0.7\textwidth]{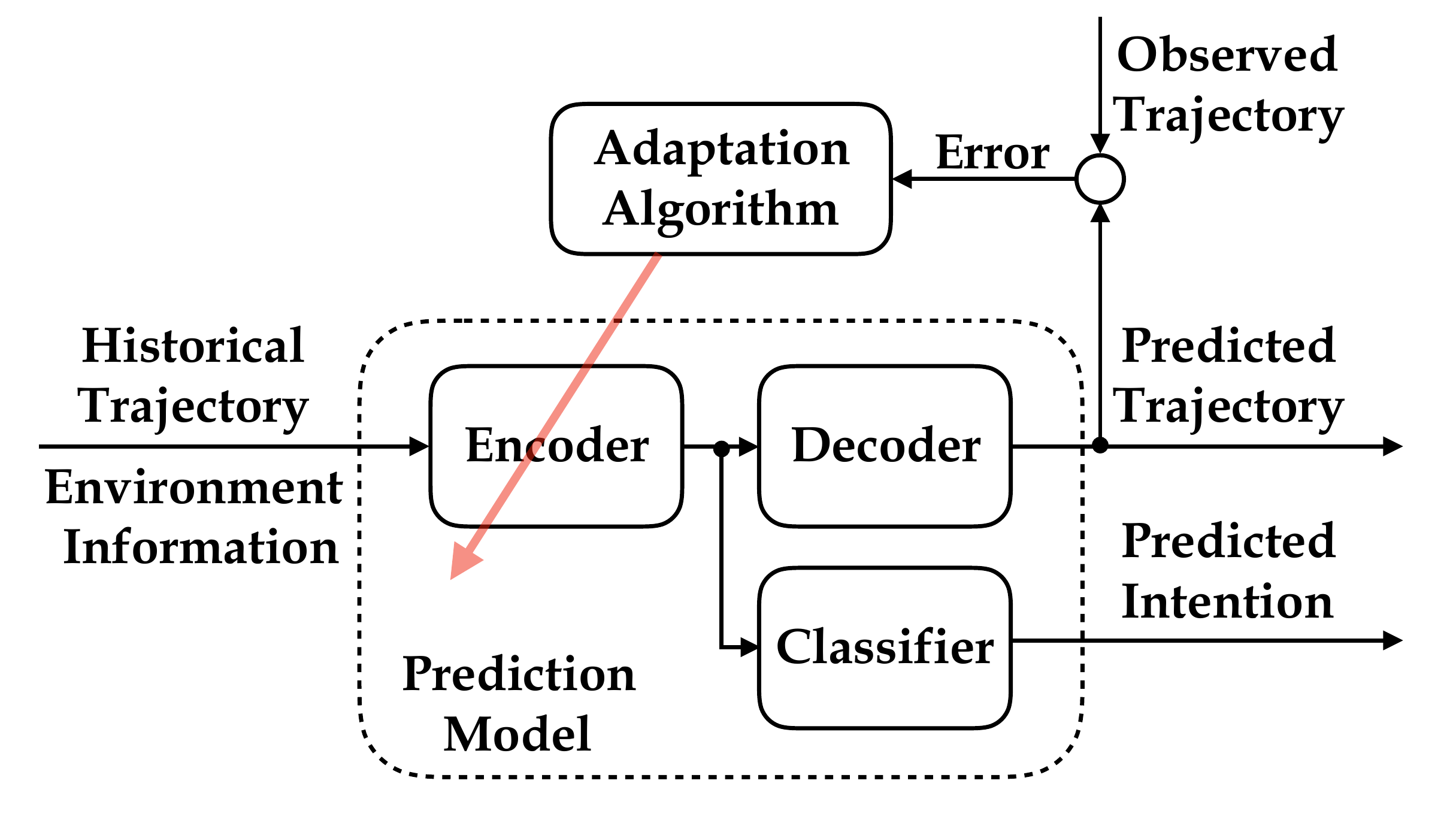}
    \caption{The online adaptation framework for a multi-task model.}
    \label{fig:adapt_frame} 
\vspace{-5pt}
\end{figure}

In online adaptation of multi-task learning, the  adaptation algorithm updates the prediction model only considering the error measured between the predicted trajectory and the ground truth trajectory. The ground truth intention is not available for online adaptation since it is not directly observable. \Cref{fig:adapt_frame} illustrates an online adaptation theme which only adapts the encoder's parameter.    

\begin{figure}[htbp]
\centering
    \includegraphics[width=0.9\textwidth]{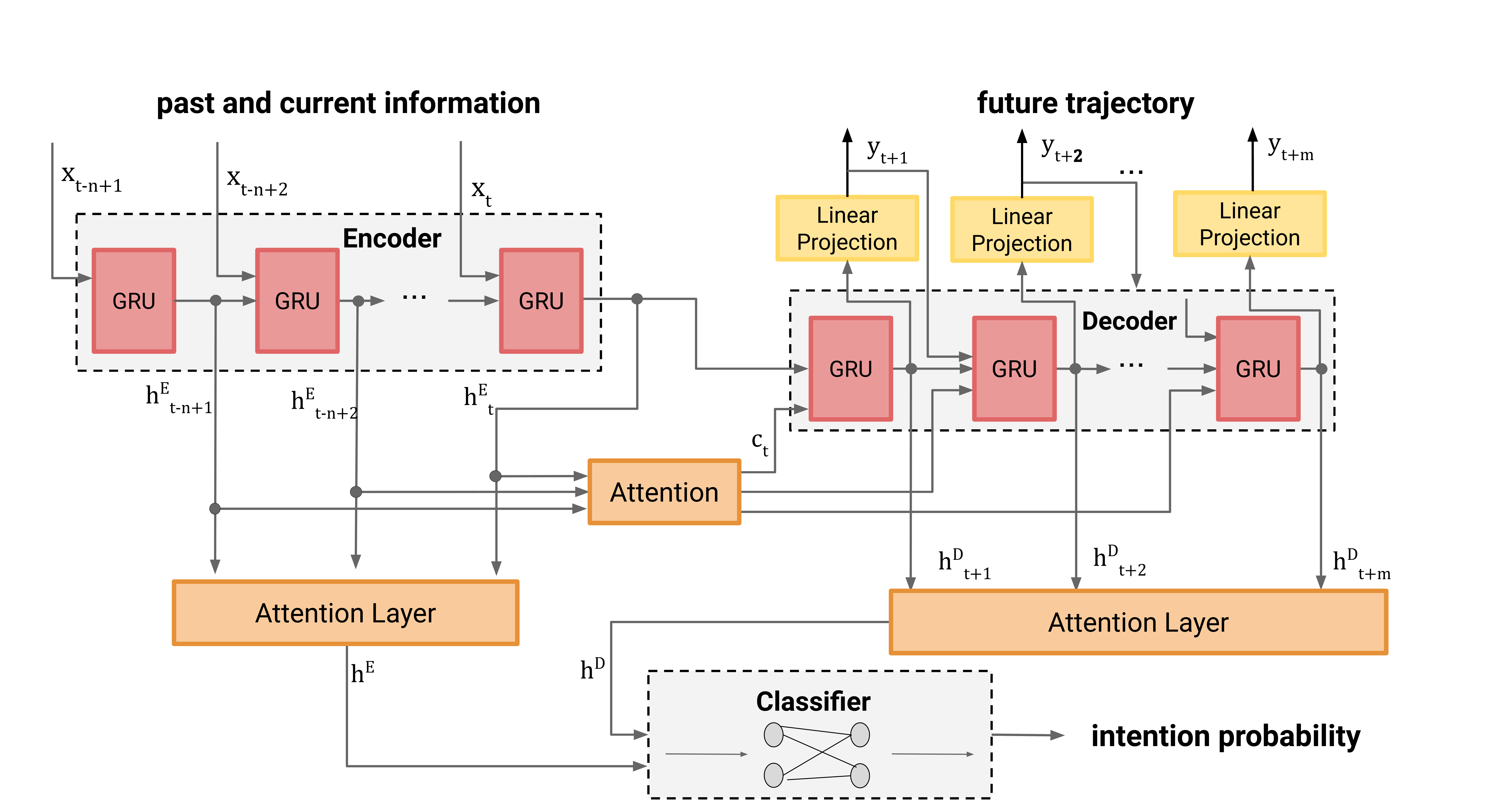}
    \caption{The neural network architecture for the RNN-based multi-task prediction model.}
    \label{fig:nn}
\vspace{-10pt}
\end{figure}

\subsubsection{Neural network architecture \label{sec:model}}

We construct Recurrent Neural Network \cite{salehinejad2017recent} (RNN) based architectures in our experiments to evaluate the  effectiveness of MEKF$_\lambda$ and MEKF$_{\text{EMA-DME}}$, as shown in \cref{fig:nn}. The neural networks follow the encoder-decoder-classifier structure for simultaneous intention and trajectory prediction as shown in \cref{fig:adapt_frame}. Trajectory prediction is based on encoder-decoder \cite{sutskever2014sequence}  structure  and intention prediction is based on encoder-classifier structure. Both the encoder and the decoder are composed of single layer Gated Recurrent Units (GRU's) \cite{DBLP:journals/corr/ChoMBB14} and the classifier is composed of two-layer FC neural networks. In order to improve the performance of offline trained model, an attention mechanism~\cite{bahdanau2014neural} is applied to the output vectors of the encoder.

\subsubsection{Dataset}
We used Mocap dataset and NGSIM dataset in our experiment. In each datset, we randomly split the dataset as 80\% offline training, 10\% offline validation and  10\% testing according to different trials.

\begin{enumerate}
 \vspace{-5pt}
 \setlength\itemsep{-0.2em}
\item Mocap dataset. This is a human-motion capture dataset collected by researchers from CMU\footnote{\href{http://mocap.cs.cmu.edu/motcat.php}{http://mocap.cs.cmu.edu/motcat.php}}. We chose the wrist trajectories of three actions (walking, running, and jumping) of all subjects in the Mocap datasets\footnote{We didn't perform full body motion prediction, since it requires special design of the neural network model to encode the geometric constraints, which is out of the scope of this paper.}. The intentions are identified with the labeled actions. There are 543 trials for all three actions.  
\item US 101 human driving data from Next Generation SIMulation (NGSIM) dataset. It is a widely used benchmark dataset for autonomous driving \cite{colyar2007us}. We extract three actions from the dataset, which are driving with constant speed, acceleration, and deceleration respectively\footnote{ In our experiment, value of acceleration $a > {\rm 0.5~m/s^2}$ was denoted as acceleration, and  value of acceleration $a < {\rm - 0.5~m/s^2}$ was denoted as deceleration.}. At time step $t$, if the vehicle will accelerate (or decelerate) in the next three seconds($t \sim t + 3 \rm{s}$) , we label the intention as acceleration (or deceleration) at time step $t$. Otherwise, we label it as constant speed. In our experiment, we used a subset of the dataset which contains 100 trials for all three actions.
 \vspace{-5pt}
\end{enumerate}

\subsubsection{Evaluation metrics}
We used accuracy to evaluate the intention prediction and average mean square error (MSE) for the trajectory prediction. The average MSE is computed as,
\begin{equation} 
 \vspace{-5pt}
{\rm MSE} =  \frac{1}{T}\sum_{t=1}^{T}{{\rm MSE}_t},~
{\rm MSE}_t = \frac{1}{m} \| \mat{Y}_t - \mat{\hat{Y}}_t \|_2,
 \vspace{-5pt}
\end{equation}
Where $T$ is total number of timesteps in the testing set. To maintain similar orders of magnitude on different datasets, we used  dm$^2$ unit for the trajectory in  Mocap dataset, and used m$^2$ unit for the trajectory in  NGSIM dataset. 

\subsubsection{Offline training}
Before online adaptation, the prediction models are trained offline. We used an Adam optimizer with a 128 batch size and a 0.01 learning rate. For the  Mocap dataset, past $n = 20$ steps input information was used to predict the trajectories of the future $m = 10$ steps and the intention. We used a concatenation of  raw trajectory and speed as the input information. For the NGSIM dataset, past $n = 20$ steps input information was used to predict the trajectories of the future $m = 50$ steps and the intention.  We used a concatenation of raw trajectories and extracted features as input information. The extracted features were similar to the features used in the parameter sharing generative adversarial imitation learning \cite{bhattacharyya2018multi}. \Cref{tab:offline} shows the prediction performance after offline learning. In experiments of the adaptation, we studied the performance of various  adaptation algorithms on hidden weights of encoder of these offline-trained models (with online adaptation on testing set).

\begin{table}[htbp]
\vspace{-5pt}
\centering
\caption{Offline training performance of the  model}
\label{tab:offline}
\begin{tabular}{l|l|l}
\hline \hline
 Metrics & CMU Mocap dataset & NGSIM dataset \\ \hline
accuracy & 0.984 & 0.951 \\
 MSE & 3.271 (dm$^2$) & 2.559 (m$^2$) \\ \hline \hline
\end{tabular}
 \vspace{-5pt}
\end{table}

\subsection{Additional experiment} \label{ap:dme}


In order to demonstrate the effectiveness of the proposed discrimination criterion in dynamic multi epoch update strategy, we design  the following experiment on the NGSIM dataset. We compared three different criteria  for DME. 

\begin{enumerate}
 \vspace{-5pt}
\setlength\itemsep{-0.2em}
\item the proposed criterion as discussed in \cref{sec:multi_epoch}. In particular, we set  $\xi_1$ as the $50\%$ quantile value of the errors , and  $\xi_2$ as the $99.9\%$ quantile value of the errors. Under the error spectrum, the first $50\%$ are "easy" samples, the middle $50\%$ to $99.9\%$ are "hard" samples, and the last $0.1\%$ are "anomaly" samples.
\item fixed criterion: we set $\kappa_t = 2$ for all samples. That means, we run fixed 2-epoch update strategy and use each sample twice.
\item random criterion: for each sample, we set $\kappa_t=1$ with the probability of $50\%$, set $\kappa_t=2$ with the probability of $49.9\%$, and set $\kappa_t=0$ with the probability of $0.1\%$. That means, the random criterion has same "easy" and "hard" ratio as the  proposed criterion, but distinguishing "easy", "hard" and "anomaly" samples randomly.
 \vspace{-5pt}
\end{enumerate}

\begin{table}[htbp]
 \vspace{-5pt}
\centering
\caption{Performance of different criterion for dynamic multi-epoch update strategy.}
\label{tab:criterion}
\begin{tabular}{l|l|l|l|l}
\hline \hline
  & w/o  DME  & fixed criterion & random criterion & proposed criterion \\ \hline
accuracy & 0.956  & 0.957 & 0.957 & \textbf{0.958}\\
MSE~(m$^2$) & 2.157  & 2.136 & 2.140 & \textbf{2.122}\\ 
\hline \hline
\end{tabular}
 \vspace{-5pt}
\end{table}

The results in \cref{tab:criterion} show that: the proposed criterion outperforms other criteria, which justifies the effectiveness of the proposed error-based criterion. Nonetheless, we will investigate more reasonable and effective criterion for dynamic multi-epoch update in the future.   

\end{document}